\documentclass{article}

\usepackage[preprint]{nips_2018}

\usepackage[utf8]{inputenc} 
\usepackage[T1]{fontenc}    
\usepackage{hyperref}       
\usepackage{url}            
\usepackage{booktabs}       
\usepackage{amsfonts}       
\usepackage{nicefrac}       
\usepackage{microtype}      
\usepackage{amsmath}
\usepackage{amssymb}
\usepackage{graphicx}
\usepackage{subcaption}
\usepackage{amsthm}
\usepackage[english]{babel}
\usepackage[toc,page]{appendix}
\usepackage{thm-restate}
\usepackage{cleveref}

\newtheorem{theorem}{Theorem}
\newtheorem{lemma}{Lemma}
\newtheorem{definition}{Definition}
\newtheorem{assumption}{Assumption}
\newcommand{\norm}[1]{\left\lVert#1\right\rVert}

\usepackage[colorinlistoftodos,bordercolor=orange,backgroundcolor=orange!20,linecolor=orange,textsize=scriptsize]{todonotes}

\title{On the Decision Boundary of Deep Neural Networks}

\author{Anonymous authors}
\author{
  Yu Li \\
  KAUST \\
  CEMSE\\
  \texttt{yu.li@kaust.edu.sa} \\
  \And
  Lizhong Ding \\
  KAUST \\
  CEMSE\\
  \texttt{lizhong.ding@kaust.edu.sa} \\
  \And
  Xin Gao \\
  KAUST \\
  CEMSE\\
  \texttt{xin.gao@kaust.edu.sa} \\
}

\begin{document}

\maketitle

\begin{abstract}
While deep learning models and techniques have achieved great empirical success, our understanding of the source of success in many aspects remains very limited. In an attempt to bridge the gap, we investigate the decision boundary of a production deep learning architecture with weak assumptions on both the training data and the model. We demonstrate, both theoretically and empirically, that the last weight layer of a neural network converges to a linear SVM trained on the output of the last hidden layer, for both the binary case and the multi-class case with the commonly used cross-entropy loss. Furthermore, we show empirically that training a neural network as a whole, instead of only fine-tuning the last weight layer, may result in better bias constant for the last weight layer, which is important for generalization. In addition to facilitating the understanding of deep learning, our result can be helpful for solving a broad range of practical problems of deep learning, such as catastrophic forgetting and adversarial attacking. 
\end{abstract}
\section{Introduction}
\label{sec:introduction}
In recent years, deep learning has achieved impressive success in various fields \cite{RN353}. Not only has it boosted the performance of the state-of-the-art methods in various areas, such as computer vision \cite{RN349} and natural language processing \cite{RN401}, it has also enabled machines to achieve human level intelligence in specific tasks \cite{RN400}. Despite its great empirical success, deep learning is often criticized for being used as a black box \cite{RN402}, which refers to the well-known gap between its empirical power and the theoretical understanding of it \cite{RN387}. 

As suggested by \cite{RN387}, a satisfactory theoretical understanding of deep learning should cover three aspects: 1) \textit{representation power}, 2) \textit{optimization characteristics}, and 3) \textit{generalization property}. The representation power of deep learning has been extensively and rigorously discussed in \cite{RN385}. In terms of the second aspect, that is, the convergence analysis of stochastic gradient descent (SGD) and the property of the minima obtained, numerous recent studies have endued promising answers \cite{RN355,RN388,RN351,RN364,RN361,RN367,RN403,RN360}. For example, \cite{RN355} proves the conjecture of \cite{RN342}, extending the result to deep nonlinear neural networks and showing the nonexistence of poor local minima. \cite{RN358} also shows that all local minima are globally optimal, given reasonable assumptions. \cite{RN403,RN360} prove the convergence of SGD given assumptions of the input distribution. 

As for the generalization mystery, the studies are still in the early stage. Through systematic experiments, \cite{RN385} suggests that although the explicit regularization, such as weight decay and dropout, may be helpful, the implicit regularization of SGD may be the key for generalization. Following that direction, \cite{RN360} provides the generalization guarantee for over-parameterized networks on linearly separable data, which are trained by SGD. \cite{RN359} shows that, for linearly separable data, gradient descent (GD) on an unregularized logistic regression problem results in the max-margin (hard margin SVM) solution. On the other hand, \cite{RN398,RN399} try to demystify the generalization property via deriving the generalization bounds.

\begin{figure}[t!]
  \centerline{\includegraphics[width = 90mm]{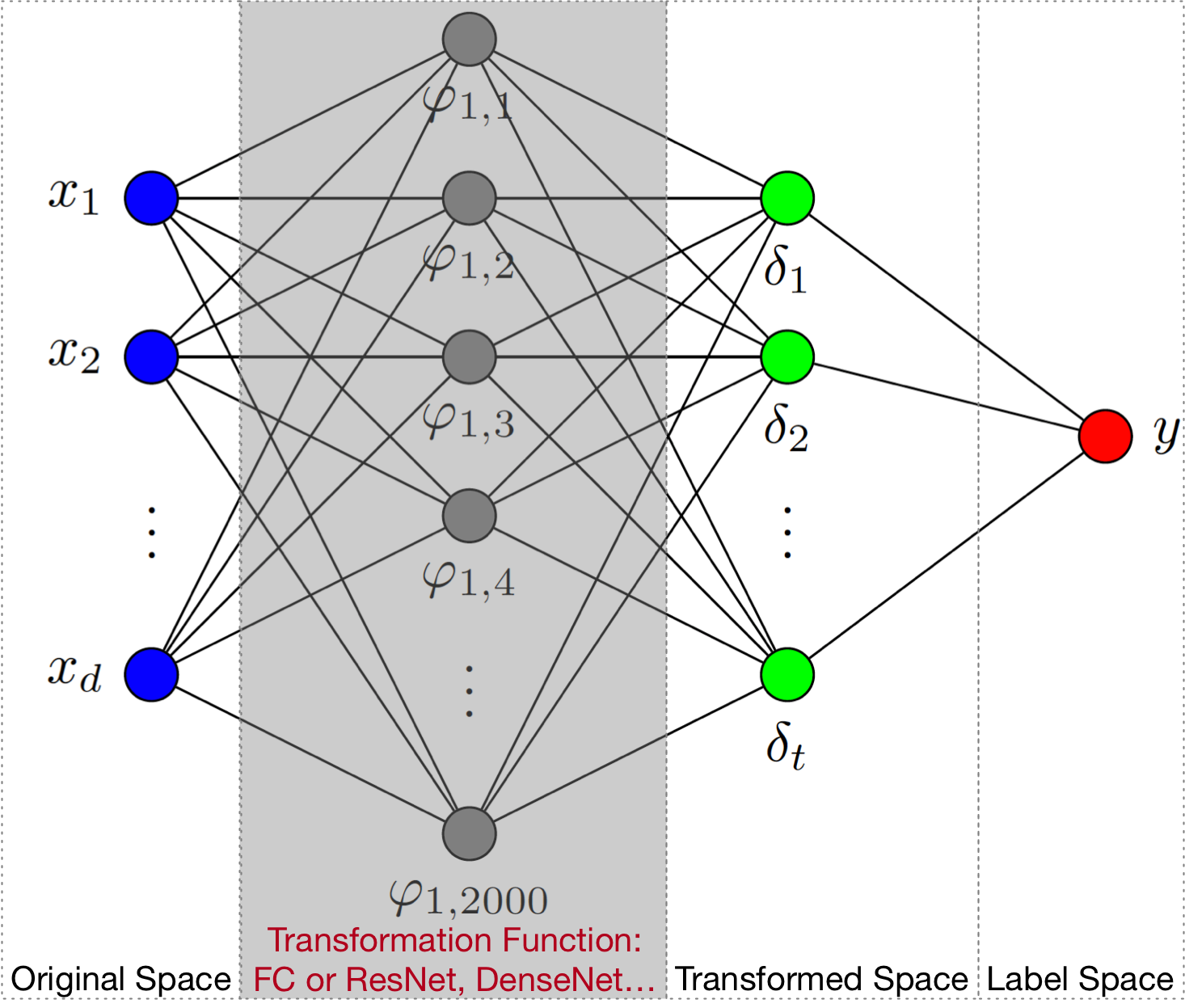}}
  \caption{Network architecture. We do not oversimplify the network, with the only assumption being that the last hidden layer and the output layer are fully connected, which is practical and the common case. The transformation function can be any kind of deep learning architecture, including the legend fully connected layers or the commonly used ResNet or DenseNet \textit{et al}. Here we show the fully connected layer for simplicity.}
  \label{fig:network}
\end{figure}

In this paper, we follow the direction of \cite{RN385,RN360,RN359}, investigating the implicit bias of GD and SGD. Unlike the previous studies, we do not oversimplify the model architecture. In fact, the architecture, which is shown in Fig. \ref{fig:network}, is a productive one, which can reach the state-of-the-art performance on CIFAR-10 if we use DenseNet \cite{RN407} as the transformation function. 
Moreover, we have little requirement for the input data distribution, only assuming that the loss converges to zero. In the Main Result section and Experiments section, we show that the direction of the neural network's last weight layer converges to that of the SVM solution trained on the transformed data in the transformed space both theoretically and empirically. In addition, we also show that the decision boundary of the last layer is closer to the SVM decision boundary if we train the whole network, instead of only fine-tuning the last layer, in the Experiments part. We extend our result to multi-class classification problem with cross-entropy loss, which is the most common scenario in practice, on the MNIST and CIFAR10. Our study bridges the gap between the purely theoretical side, which investigates the over-simplified models and has strict requirements for the input distribution, and the practical usage of complex deep learning models. In practice, people usually owe the superior performance of deep learning to the model's ability of learning representation and classifier simultaneously. We demystify the relationship between the learned representation and the classifier, and characterize the learned classifier in particular.

\section{Problem Formulation}
\label{sec:problem_formulation}
Unlike the setting of previous studies \cite{RN359,RN360,RN403}, which assume the training data is linearly separable or follows a certain distribution, we do not have such a requirement. Formally, for binary classification, we consider a dataset $\{\textbf{x}_n, y_n\}_{n=1}^N$, with $\textbf{x}_n \in \mathbb{R}^d$, and binary labels $y_n \in \{-1, 1\}$. We use $\textbf{X} \in \mathbb{R}^{d*N}$ to denote the data matrix. For multi-class classification, we have $y_n \in [K]:=\{1,2,\dots,K\}$ and $K$ is the number of classes. 

Regarding the neural network model, we do not restrict to any specific the architecture neither. Consider a neural network with the architecture shown in Fig.~\ref{fig:network}, which is basically a production network with practical usage. We divide the neural network into four components. The original space and label space are the training interface. The transformation function combined with the transformed space (the output of the last hidden layer) is one of the reasons why the deep learning's performance is being continuously improved. For the sake of analysis, we take the transformed space as an independent component which is fully connected with the label space. Formally, we denote the output of the last hidden layer on example $\textbf{x}_n$ as $\boldsymbol{\delta}_n$, with $\boldsymbol{\delta}_n \in \mathbb{R}^t$.

We denote the entire parameter set of the network as $\theta$. The network defines a function $f(\textbf{x}; \theta): \mathbb{R}^d \rightarrow \{-1, 1\}$ for the binary case. The transformation function is $\boldsymbol{\delta}_n = h(\textbf{x}_n; \boldsymbol\phi)$, where $\boldsymbol\phi$ is the parameter set of the transformation function. Notice that from $\boldsymbol{\delta}_n$ to the final output, the last weight layer defines a linear transformation, which has the following form:
\begin{equation}
\begin{gathered}
g(\boldsymbol{\delta}_n; \textbf{W}) = \textbf{W}\boldsymbol{\delta}_n,
\label{eq:last_layer_function}
\end{gathered}
\end{equation}
where $\textbf{W} \in \mathbb{R}^{t*k}$ is the weight vector of the last layer (notice that for the binary case, $k=1$). We use $W_i \in \mathbb{R}^{t*1}$ to denote the $i$-th row of it. So, we have $\theta = (\boldsymbol\phi,\textbf{W})$. 


In general, the empirical loss over the training dataset has the following form:
\begin{equation}
\begin{gathered}
L(\theta) = \sum_{n=1}^N l(f(\textbf{x}_n; \theta), y_n),
\label{eq:loss}
\end{gathered}
\end{equation}
where $l$ is the specified loss function (e.g., exponential loss, cross-entropy, \dots). For example, with the exponential loss, $l(t,y_n) = e^{-y_n t}$, the empirical risk is given by
\begin{equation}
\begin{gathered}
L_{exp}(\theta) = \sum_{n=1}^N e^{-y_n f(\textbf{x}_n; \theta)}; \\
L_{exp}(\textbf{W}, \boldsymbol\phi) = \sum_{n=1}^Ne^{-y_n \textbf{W}\boldsymbol{\delta}_n(\textbf{x}_n;\boldsymbol\phi)}, 
\label{eq:exp_loss}
\end{gathered}
\end{equation}
where the second expression emphasizes the last weight layer.

For multi-class classification, the commonly used loss function is cross-entropy loss:
\begin{equation}
\begin{gathered}
L_{cross-entropy}(\textbf{W}, \boldsymbol\phi) = -\sum_{n=1}^N \log \left(\frac{\exp(W_{y_n}\boldsymbol{\delta}_n(\textbf{x}_n;\boldsymbol\phi))}{\sum_{l=1}^K \exp(W_l\boldsymbol{\delta}_n(\textbf{x}_n;\boldsymbol\phi))}\right),
\label{eq:cross-entropy}
\end{gathered}
\end{equation}
where 
$W_{l}$ is the $l$-th component of $\textbf{W}$, which is the weight for a certain class $l$; $W_{y_n}$ is the component of $\textbf{W}$ for the class represented by $y_n$.


The goal of performing optimization is to find:
\begin{gather*}
\arg\min_{\theta} L(\theta).
\label{eq:optimization}
\end{gather*}
In the following, we focus on minimizing Equation (\ref{eq:exp_loss}) using GD algorithm with a constant learning rate $\eta$ for the binary case and Equation (\ref{eq:cross-entropy}) for the multi-class case. At iteration $t$, the update rule has the following form: 
\begin{equation}
\begin{gathered}
\theta_{t} = \theta_{t-1} - \eta\nabla L(\theta_{t-1}).
\label{eq:gd}
\end{gathered}
\end{equation}

\section{Main Result} 
\label{sec:main_result}
In this section, we start with the result in \cite{RN359} for linearly separable data in logistic regression and then obtain the result for the neural network in Fig. \ref{fig:network}. Finally, we extend the result from the binary case to the multi-class case.

In \cite{RN359}, the authors investigate the following problem. 
\begin{definition}
For a logistic regression problem, whose weight vector is $\textbf{w} \in \mathbb{R}^d$, the loss has the following form:
\begin{gather*}
L_{logistic}(\textbf{w}) = \sum_{n=1}^Nl(y_n\textbf{w}^\top\textbf{x}_n).
\end{gather*}
For this binary case, assuming all the labels are positive $\forall n: y_n=1$ (we can re-define $y_n\textbf{x}_n$ as $\textbf{x}_n$), we have the GD update for that loss function at iteration $t$ having the following form:
\begin{gather*}
\textbf{w}_{t} = \textbf{w}_{t-1} - \eta\nabla L_{logistic}(\textbf{w}_{t-1}) \\
= \textbf{w}_{t-1} - \eta \sum_{n=1}^N l^{'}(\textbf{w}_{t-1}^\top\textbf{x}_n)\textbf{x}_n.
\end{gather*}
\label{def:logistic}

\end{definition}

The authors show that $\textbf{w}_t$ finally diverges \cite{RN359}:
\begin{lemma}
Let $\textbf{w}_{t}$ be the iterates of gradient descent in Definition \ref{def:logistic} with $\eta < 2\beta^{-1}\sigma^{-2}_{\max}(\textbf{X})$, 
where $\beta$ is the smoothness of l and $\sigma_{\max}(\textbf{X})$ is the maximal singular value of the data matrix $\textbf{X} \in \mathbb{R}^{d*N}$ and any starting point $\textbf{w}_{0}$. For linearly separable data and $\beta$-smooth decreasing loss function, we have: (1) $\lim_{t \rightarrow \infty} L_{logistic}(\textbf{w}_t)=0$, (2) $\lim_{t \rightarrow \infty}\norm{\textbf{w}_{t}}=\infty$ and (3) $\forall n: \lim_{t \rightarrow \infty} \textbf{w}_t^\top\textbf{x}_n=\infty$.
\label{lemma:w_diverge}
\end{lemma}
But the direction of the above solution converges to that of the hard margin SVM solution \cite{RN359}.
\begin{lemma}
For any dataset which is linearly separable, any $\beta$-smooth decreasing loss function with an exponential tail (the loss function tail is bounded by two exponential functions), any step size $\eta < 2\beta^{-1}\sigma_{\max}^{-2}(\textbf{X})$ and any starting point $\textbf{w}_0$, the gradient descent iterations will behave as:
\begin{equation}
\begin{gathered}
\textbf{w}_t = \hat{\textbf{w}}\log t+ \boldsymbol{\rho}_t,
\end{gathered}
\end{equation}
where $\hat{\textbf{w}}$ is the $L_2$ max margin vector:
\begin{equation}
\begin{gathered}
\hat{\textbf{w}} = \arg\min_{\textbf{w} \in \mathbb{R}^{d}} \norm{\textbf{w}}^2 \quad \\
\text{subject to} \quad \textbf{w}^\top\textbf{x}_n \geq 1,
\end{gathered}
\end{equation}
and the residual grows at most as $\norm{\boldsymbol{\rho}_t}$ = $O(\log\log(t))$, and so
\begin{gather*}
\lim_{t \rightarrow \infty} \frac{\textbf{w}_t}{\norm{\textbf{w}_t}} = \frac{\hat{\textbf{w}}}{\norm{\hat{\textbf{w}}}}.
\end{gather*}
Furthermore, except for measuring zero, the residual $\boldsymbol{\rho}_t$ is bounded.
\label{lemma:logistic_converge}
\end{lemma}

As for our problem, we have the following assumption:
\begin{assumption}
The loss in Equation (\ref{eq:loss}) converges to zero: $\lim_{t \rightarrow \infty} L(\theta_t)=0$.
\label{ass:loss_to_zero}
\end{assumption}
This assumption is a reasonable assumption. It could be satisfied as long as the data is linearly or non-linearly separable, with no wrongly labeled data points and the model has enough capacity, which is usually the case for deep learning models. 
Based on Assumption \ref{ass:loss_to_zero}, we have the following lemma:
\begin{lemma}
Under Assumption \ref{ass:loss_to_zero}, for the neural network with architecture as in Fig. \ref{fig:network}, even if the dataset $\{\textbf{x}_n, y_n\}_{n=1}^N$ is not linearly separable, the transformed dataset $\{\boldsymbol{\delta}_n, y_n\}_{n=1}^N$ is linearly separable: $\exists \textbf{W}^*$ such that $\forall n: y_n\textbf{W}^*\boldsymbol{\delta}_n>0$.
\label{lemma:linear_sep}
\end{lemma}
In fact, since the last weight layer is a linear transformation, if $\{\boldsymbol{\delta}_n, y_n\}_{n=1}^N$ is not linearly separable, the classification error can never reach zero, let alone the loss.
Following Definition \ref{def:logistic}, let us re-define $y_n\boldsymbol{\delta}_n$ as $\boldsymbol{\delta}_n$,
Based on Lemma \ref{lemma:logistic_converge} and Lemma \ref{lemma:linear_sep}, we obtain the first main result: 
\begin{theorem}
For any neural network for binary classification, any $\beta$-smooth decreasing loss function with an exponential tail, small enough step size $\eta < 2\beta^{-1}\sigma_{\max}^{-2}(\textbf{X})$ and any start point $\textbf{W}_0$, as long as $\lim_{t \rightarrow \infty} L(\theta_t)=0$, the direction of the neural network's last weight layer converges:
\begin{equation}
\begin{gathered}
\lim_{t \rightarrow \infty} \frac{\textbf{W}_t}{\norm{\textbf{W}_t}} = \frac{\hat{\textbf{W}}}{\norm{\hat{\textbf{W}}}},
\end{gathered}
\end{equation}
where $\hat{\textbf{W}}$ is the $L_2$ max margin vector:
\begin{gather*}
\hat{\textbf{W}} = \arg\min_{\textbf{W} \in \mathbb{R}^{t*1}} \norm{\textbf{W}}^2 \quad \\
\text{subject to} \quad \textbf{W}\boldsymbol{\delta}_n \geq 1,
\end{gather*}
in which $\boldsymbol{\delta}_n$ is the re-defined input of the last weight layer.
\label{the:last_layer_exp}
\end{theorem}
It is true that the convergence of the transformation function can also affect the last layer decision boundary. However, since the loss converges to zero, the variance of the transformation function is bounded after long enough training time, which makes the theorem hold.

As for the multi-class classification problem, we have the following lemma from \cite{RN359}:
\begin{lemma}
For a logistic regression problem in which we learn a predictor $\textbf{w}_k$ for each class $k \in [K]$ in a linearly separable multi-class dataset, any starting point $\textbf{w}_{k,0}$ and any small enough step size, under most circumstances (i.e., except for a measure zero), the iterates of gradient descent on the cross-entropy loss will behave as:
\begin{equation}
\begin{gathered}
\textbf{w}_{k,t} = \hat{\textbf{w}}_k\log(t) + \boldsymbol{\rho}_{k,t},
\end{gathered}
\end{equation}
where the residual $\boldsymbol{\rho}_{k,t}$ is bounded and $\hat{\textbf{w}}_k$ is the solution of the K-class SVM:
\begin{equation}
\begin{gathered}
\arg\min_{\textbf{w}_1,...,\textbf{w}_k}\sum_{k=1}^K\norm{\textbf{w}_k}^2 \quad \\
\text{subject to} \\
\quad \forall n, \forall k \neq y_n: \textbf{w}_{y_n}^\top\textbf{x}_n \geq \textbf{w}_k^\top\textbf{x}_n +1.
\end{gathered}
\end{equation}
\label{lemma:ce_converge}
\end{lemma}

\begin{figure*}[ht!]
  \centerline{\includegraphics[width = 180mm]{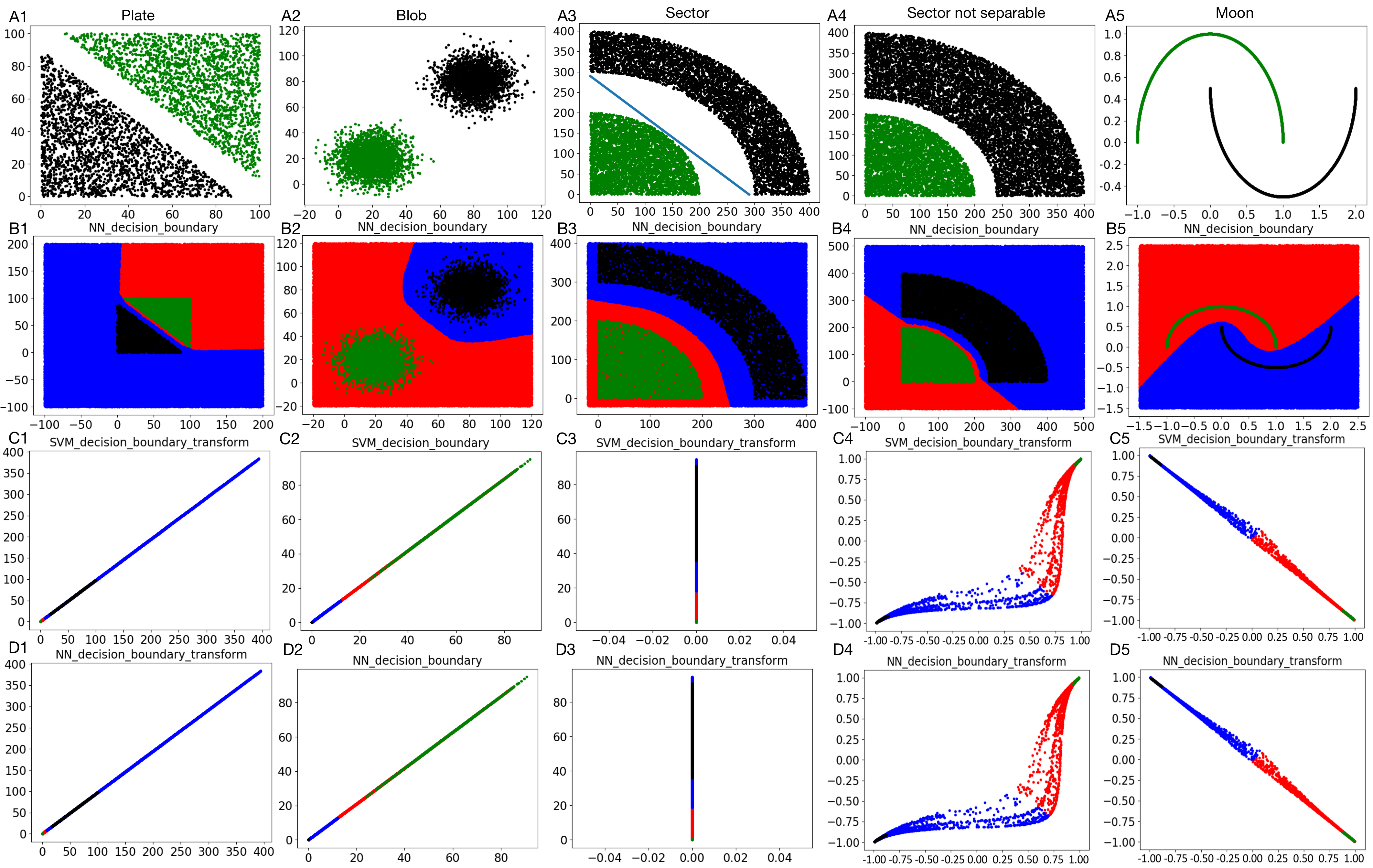}}
  \caption{Results on simulated datasets. The five columns are five datasets. The first row is the training datasets in the original input space. In the last three rows, red points are random testing points classified with the same label as the green training data points and blue points are random testing points classified with the same label as the black training data points. The interface between red dots and blue dots is the decision boundary. The second row figures show the decision boundary of a trained neural network in the original space. The third and the fourth rows are in the transformed space. The third row shows the decision boundary of linear SVM trained with the transformed data in the transformed space. The last row shows the decision boundary of the neural network's last weight layer. }
  \label{fig:simulated_data_result}
\end{figure*}

Similar to Theorem \ref{the:last_layer_exp}, we can derive the following result for the multi-class case with cross-entropy loss.
\begin{theorem}
For any neural network, small enough step size $\eta$ and any starting point $\textbf{W}_0$, as long as the dataset makes $\lim_{t \rightarrow \infty} L(\theta_t)=0$, the iterates of gradient descent on the cross-entropy loss of the last weight layer $\textbf{W}$ will behave as:
\begin{equation}
\begin{gathered}
W_{k,t} = \hat{W}_k\log(t) + \boldsymbol{\rho}_{k,t},
\end{gathered}
\end{equation}
where the residual $\boldsymbol{\rho}_{k,t}$ is bounded, $W_{k,t}$ is the weight for class $k$ at iteration $t$ and $\hat{{W}}_k$ is the solution of the $K$-class SVM:

\begin{gather*}
\arg\min_{W_1,...,W_k}\sum_{k=1}^K\norm{W_k}^2 \quad \\
\text{subject to}  \\
\quad \forall n, \forall k \neq y_n: W_{y_n}^\top\boldsymbol{\delta}_n \geq W_k^\top\boldsymbol{\delta}_n +1,
\end{gather*}

and so:
\begin{equation}
\begin{gathered}
\lim_{t \rightarrow \infty} \frac{W_{t,k}}{\norm{W_{t,k}}} = \frac{\hat{W}_k}{\norm{\hat{W}_k}}.
\end{gathered}
\end{equation}

\label{the:last_layer_ce}
\end{theorem}


\begin{figure*}[t!]
  \centerline{\includegraphics[width = 140mm]{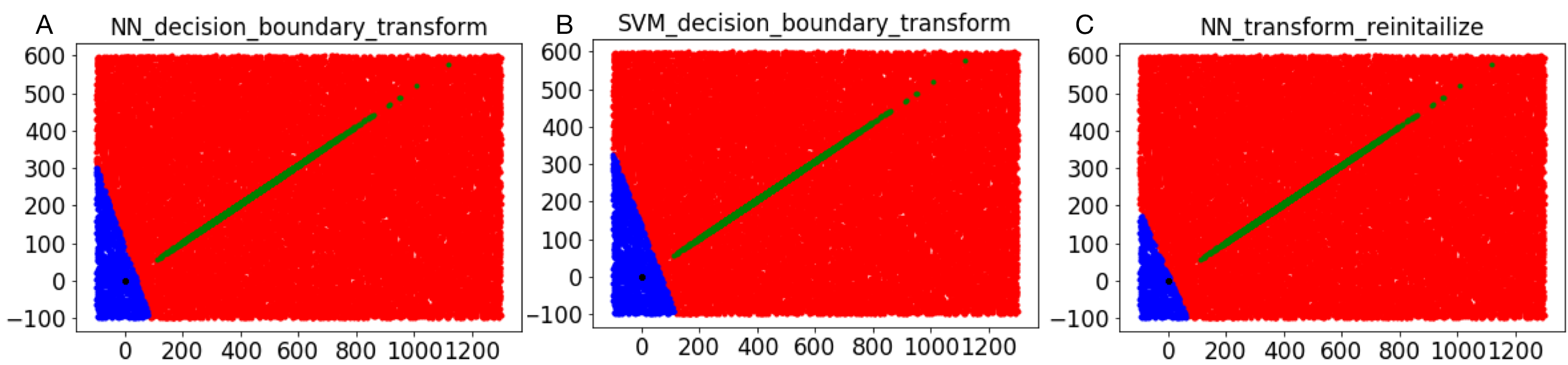}}
  \caption{MNIST binary classification decision boundary. We randomly chose two classes from the MNIST dataset (``0'' and ``1'' for the above figures) and trained a neural network with ResNet as the transformation function in Fig. \ref{fig:network}. We set $t$ as 2 for visualization purpose. The above figures show the decision boundary of SVM trained with transformed data (B) and the last weight layer of the neural network (A) in the transformed space. After model converged, we reinitialized the last weight layer and retrained the model with the weights in the transformation function being fixed, resulting in (C). }
  \label{fig:mnist}
\end{figure*}

\section{Experiments} 
\label{sec:experiments}
\subsection{Experimental setting} 
\label{sub:experiment_setting}
There are seven datasets in our experiments, including five simulated 2D datasets and two real datasets. The five simulated datasets can be referred to Fig. \ref{fig:simulated_data_result} (A1-A5). The first three (Plate, Blob, and Sector) are linearly separable. The last two (Sector not separable and Moon) are non-linearly separable. There are 5000 points within each simulated dataset. The two real datasets are MNIST \cite{RN409} and CIFAR-10 \cite{RN410}. Since MNIST and CIFAR-10 are multi-class datasets, we randomly chose two classes out of the 10 classes for each one for the binary classification case. We used the network architecture in Fig. \ref{fig:network} for all the experiments. The only difference is the transformation function. We used a fully connected layer with 2000 nodes as the transformation function for the five simulated datasets; ResNet \cite{RN406} for MNIST; and DenseNet \cite{RN407} for CIFAR-10. For visualization purpose, we set $t$ as $2$. We used cross-entropy loss as the loss function and ReLU as the activation function. For multi-class classification problem, we set the number of nodes in the output layer the same as the number of classes. We used GD for the simulated datasets and SGD for MNIST and CIFAR-10. We turned off all the commonly used explicit regularizers, such as weight decay and dropout, for all the experiments.


\subsection{Simulated datasets} 
\label{sub:simulated_datasets}
The results are summarized in Fig. \ref{fig:simulated_data_result} (additional results can be found in the Appendices). The decision boundary of neural networks in the original input space can be referred to Fig. \ref{fig:simulated_data_result} (B1-B5). The green and black dots are the training data points. We sampled test data points uniformly across the whole space so that we can visualize the decision boundary of the trained neural networks. The blue points are the ones predicted by the model with the same label as the black training data while the red points are the ones predicted with the same label as the green training data. The curve that separates the blue points and red points can be considered as the decision boundary of the network. Although it is difficult to gain insight from the original space, as suggested by the analysis in the Main Result section, the transformed space is more interesting. Fig. \ref{fig:simulated_data_result} (D1-D5) shows the training data and testing data in the transformed space. As a comparison, we trained a linear SVM with the transformed training data and labeled the same testing data points with the SVM classifier, whose results are shown in Fig. \ref{fig:simulated_data_result} (C1-C5). As shown in the figure, the direction of the neural network's last layer decision boundary trained with GD converges to that of the linear SVM solution, which verifies Theorem \ref{the:last_layer_exp}. Furthermore, the two kinds of decision boundaries are very close to each other, not only in the direction but also in the constant bias term. We further discuss this phenomenon in the next subsection.

\subsection{MNIST binary} 
\label{sub:mnist}
After training a residual network with the MNIST data, we mapped the data into the transformed space. Within that space, we sampled test data uniformly and labeled those test data points using the last layer of the network in Fig. \ref{fig:network}, which results in the decision boundary in Fig. \ref{fig:mnist} (A). Utilizing the training data in the transformed space, we trained a linear SVM classifier and plotted out the decision boundary of that classifier in Fig. \ref{fig:mnist} (B). As shown in the figures, after mapping the data into the transformed space, the direction of the first decision boundary is very close to that of the second decision boundary, which further supports Theorem \ref{the:last_layer_exp}. 
Furthermore, with the transformation function fixed, we reinitialized the last layer and retrained the last layer, whose result is shown in Fig. \ref{fig:mnist} (C). It suggests that our result still holds. On the other hand, the original boundary obtained by training the network as a whole is closer to the SVM boundary in terms of the bias constant, which suggests the whole network training may have better initialization for the last layer and thus make the model generalize better. Notice that although we turned off dropout and the model had been trained for a very long time to make it completely fit to the training data, the trained model still has very impressive generalization property with the testing accuracy being as high as 99.7\%.

\begin{figure}[h!]
  \centerline{\includegraphics[width = 90mm]{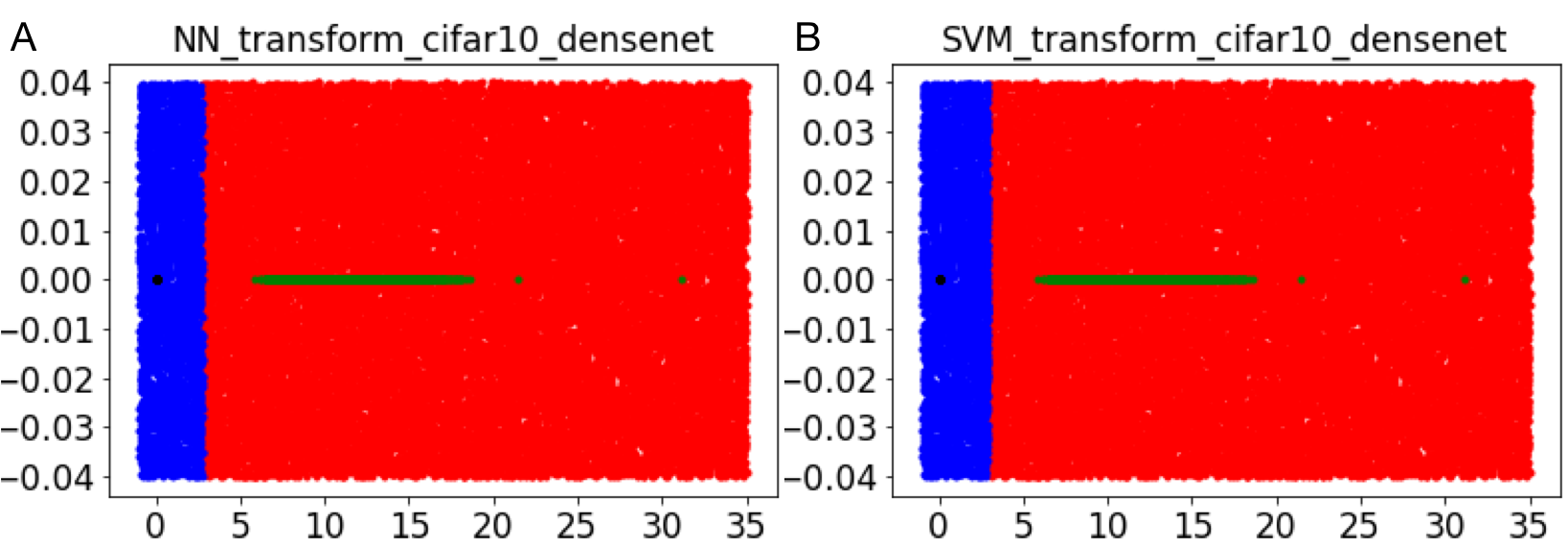}}
  \caption{CIFAR-10 binary decision boundary result. We randomly chose two classes from the CIFAR-10 dataset and trained a neural network with DenseNet as the transformation function in Fig. \ref{fig:network}, setting $t$ as 2. (A) shows the decision boundary of the last weight layer in the transformed space and (B) shows the decision boundary of the linear SVM trained with the transformed data in the transformed space. }
  \label{fig:cifar}
\end{figure}

\begin{figure*}[t!]
  \centerline{\includegraphics[width = 140mm]{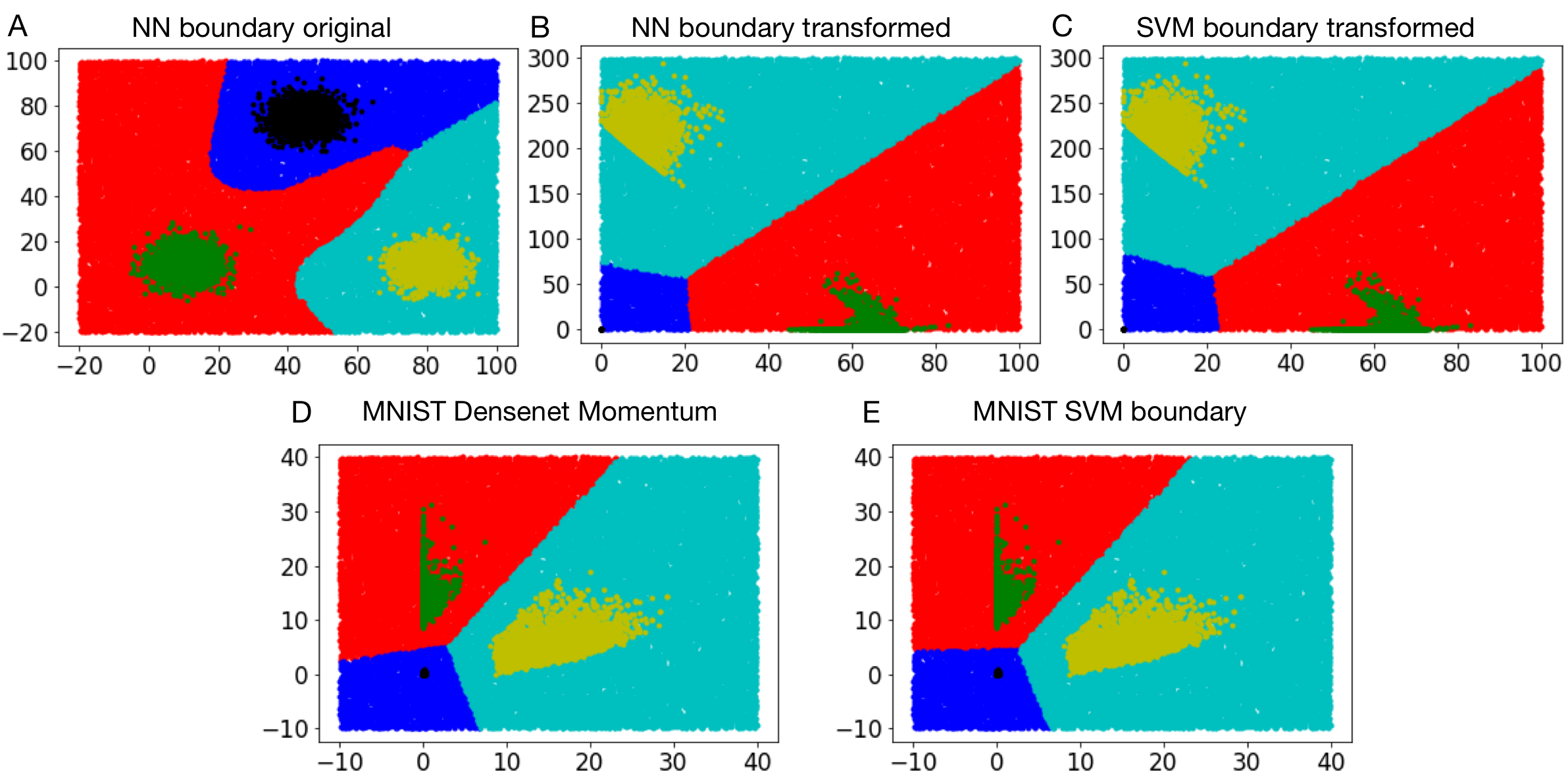}}
  \caption{The multi-class experiment result. (A,B,C) show the decision boundary results on a simulated 3-class Blob dataset. (D,E) show the 3-class MNIST (``0'', ``1'' and ``2'' for the above figures) result trained with DenseNet and Momentum.}
  \label{fig:multiclass}
\end{figure*}

\subsection{CIFAR-10 binary} 
\label{sub:cifar_10}
We trained a model with DenseNet transformation function on the CIFAR-10 dataset. The decision boundary results of this dataset could be referred to Fig. \ref{fig:cifar}. As shown in the figure, similar to the result on MNIST, the directions of those two boundaries are very close to each other, which further supports Theorem \ref{the:last_layer_exp}. Furthermore, in addition to being close in terms of direction, the neural network boundary is very close to the midpoint of the two clusters, if it does not cross the midpoint, where the SVM boundary should pass theoretically. This phenomenon is consistent with the result of the simulated datasets and the MNIST dataset, suggesting that training the whole neural network using GD or SGD may result in a decision boundary with good bias constant. In terms of the trained model's generalization property, although we turned off explicit regularizers, the model can still have 92.6\% testing accuracy for this CIFAR-10 dataset, which is within the performance range of a productive deep learning model.

\subsection{Multi-class classification} 
\label{sub:multi_class}
In practice, deep learning is usually used for multi-class classification with cross-entropy loss. We investigated the multi-class classification case in this section. We performed experiments on a simulated three class Blob dataset. The neural network decision boundary in the original space and the transformed space can be referred to Fig.~\ref{fig:multiclass} (A,B), respectively. As a comparison, the SVM decision boundary on the transformed data in the transformed space is shown in Fig.~\ref{fig:multiclass} (C). Those results, which show the decision boundary direction of the neural network last weight layer converges to that of SVM, verify Theorem \ref{the:last_layer_ce}. We also performed such experiment on the MNIST data with DenseNet transformation function. During the training, we also tried other optimizers other than just SGD, such as Momentum. The results are shown in Fig.~\ref{fig:multiclass} (D,E). From the two figures, we can find that the corresponding decision boundary directions of the neural network last layer and SVM are very close to each other. Besides, similar to the previous result, the decision boundary of neural network is very close to the midpoint between different clusters. Those experiments further support Theorem \ref{the:last_layer_ce}, which also shows that our hypothesis may be generalized to other optimizers, such as Momentum.

\begin{figure}[h!]
  \centerline{\includegraphics[width = 90mm]{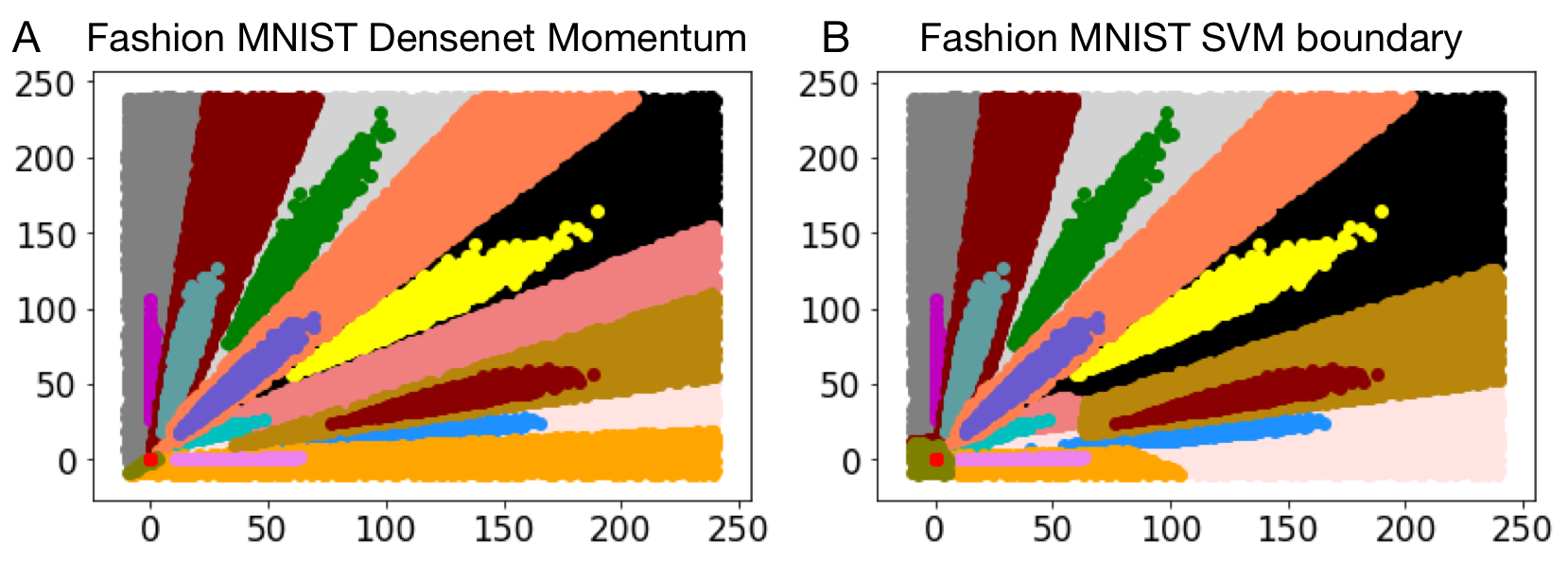}}
  \caption{The real task result. We trained a DenseNet for the 10-class Fashion MNIST classification using Momentum. After 1,000 epoches, the loss is around $6*10^{-4}$. (A) shows the decision boundary of the neural network's last layer. (B) shows the decision boundary of SVM trained with the transformed dataset.}
  \label{fig:fashion_mnist}
\end{figure}

\subsection{Real task: Fashion MNIST} 
\label{sub:real_task}
We also investigated the decision boundary of the DenseNet's last layer, which is used to perform 10-class classification on Fashion MNIST. We used the same architecture from \cite{RN407}, except for that we added an additional layer to make the last hidden layer in 2D space for visualization purpose. We turned off the commonly used techniques for improving performance, such as data augmentation and dropout. We deployed Momentum as the optimizer. After the model being trained for 1,000 epoches, the loss oscillated around $6*10^{-4}$. The testing accuracy is around 91.8\%, which is within the known performance range of the deep learning model on this dataset. We show the decision boundary comparison of the network's last layer and the multi-class linear SVM solution in Fig.~\ref{fig:fashion_mnist}. As shown in the figure, although the experiment setting is not exactly the same as the assumptions in our main result, the decision boundary of the trained neural network still worths investigating. In fact, the decision boundary shown on the up-left of Fig.~\ref{fig:fashion_mnist} (A) is very similar to that of Fig.~\ref{fig:fashion_mnist} (B). On the other hand, the transformed representation of the blue class has very complex spatial relationship with the other three classes around it, which causes the neural network get stuck in local minimum and diverge from the multi-class linear SVM solution. Although this real task does not completely fit our assumption and our result, the experiment shows that the margin theory can have the potential to explain the generalization property of deep learning.

\section{Discussion} 
\label{sec:discussion}
The result of this paper can be useful for solving several practical problems related to deep learning, such as catastrophic forgetting \cite{RN408} and the data-hungry challenge \cite{RN412}. We take these two as examples. On the other hand, we believe that investigating the transformation function would be helpful for solving adversarial attacking \cite{RN411} and studying the last layer can push out new ways of introducing uncertainty into supervised deep learning \cite{RN413}.



\subsection{Catastrophic forgetting} 
\label{sub:catastrophic_forgetting}
Catastrophic forgetting \cite{RN408}, which means the neural network does not have the ability of learning new knowledge without forgetting the learned knowledge, is one of the bottlenecks of deep learning. Recently, a rehearsal framework, called SupportNet \cite{RN404}, was proposed to deal with catastrophic forgetting when performing class incremental learning. In short, it maintains a subset of the old data, which is chosen based on the support vector information obtained by using SVM to approximate the last layer, and feeds the subset together with the new data to the model when incorporating the new classes into the model. Despite the lack of theoretical analysis in the paper, the framework works quite well in practice, even achieving nearly optimal performance on some datasets. In fact, according to Lemma \ref{lemma:w_diverge} and Theorem \ref{the:last_layer_ce}, we can write $W_{k,t} = c(t)\hat{W}_k + \boldsymbol{\rho}_{k,t}$ such that $c(t) \rightarrow \infty$ and $\boldsymbol{\rho}_{k,t}$ is bounded. The gradient of the exponential loss for $W_{k,t}$ can then be formulated as:
\begin{gather}
\begin{split}
-\nabla L_{exp}(W_{k,t}) &= \sum_{n=1}^N\exp(-W_{k,t}\boldsymbol{\delta}_n)\boldsymbol{\delta}_n \\
& =\sum^N_{n=1}\exp(-c(t)\hat{W}_k\boldsymbol{\delta}_n)\exp(-\boldsymbol{\rho}_{k,t}\boldsymbol{\delta}_n)\boldsymbol{\delta}_n,
\end{split}
\end{gather}
when the model converges and $c(t) \rightarrow \infty$ , only those data with the largest exponents, that is, $\hat{W}_k\boldsymbol{\delta}_n$ should be the smallest, will contribute to the gradients. Those samples are exactly the support vectors of the SVM trained on the transformed data, which are selected by SupportNet. Using those data for future tuning, the model is very likely to learn the same boundary for the old classes. Our results partially explains why that rehearsal method works very well in practice.

\subsection{Reducing the training data size and transfer learning} 
\label{sub:reducing_the_training_data_size_and_transfer_learning}
It is always desirable to reduce the training data size for the data-hungry deep learning method, without too much performance compromise. In practice, especially in the computer vision field, when the data size is not large enough, people usually take advantage of transfer learning \cite{yosinski2014transferable}, fine-tuning the last one or two layers of a pre-trained model with the training data. In fact, based on our result in the Main Result section and the analysis in the previous subsection, it is not data-hungry from the transformed space to the label space since only the support vector samples matter, which means the drawback property of deep learning comes from the transformation function component. The transfer learning technique, taking advantage of an existing transformation function and avoiding the data size requirement of that component, can thus learn a useful model with limited data.



\section{Conclusion} 
\label{sec:conclusion}
Bridging the gap between the theoretical research and the practical power of deep learning is a fascinating research direction. In this paper, we investigate the decision boundary of a productive deep learning architecture with weak assumption on both the training data and the model. Through comprehensive theoretical analysis and experiments, we show that the direction of the neural network's last weight layer converges to that of a linear SVM trained on the transformed data if the loss converges to zero, for both the binary case and the multi-class case with the commonly used cross-entropy loss. In addition, we show it empirically that training a neural network as a whole may result in better bias constant for the last weight layer, which is important for the generalization property of deep learning models. In addition to facilitating the understanding of deep learning and thus further improving its performance, our result can be useful for solving a broad range of practical problems in the deep learning field, such as catastrophic forgetting, reducing the data size requirement of deep learning, adversarial attacking, and introducing uncertainty into deep learning. 

\bibliographystyle{plain}
\bibliography{references}

\clearpage

\begin{appendices}

\section{Additional results of binary classfication}
\begin{figure*}[h!]
  \centerline{\includegraphics[width = 150mm]{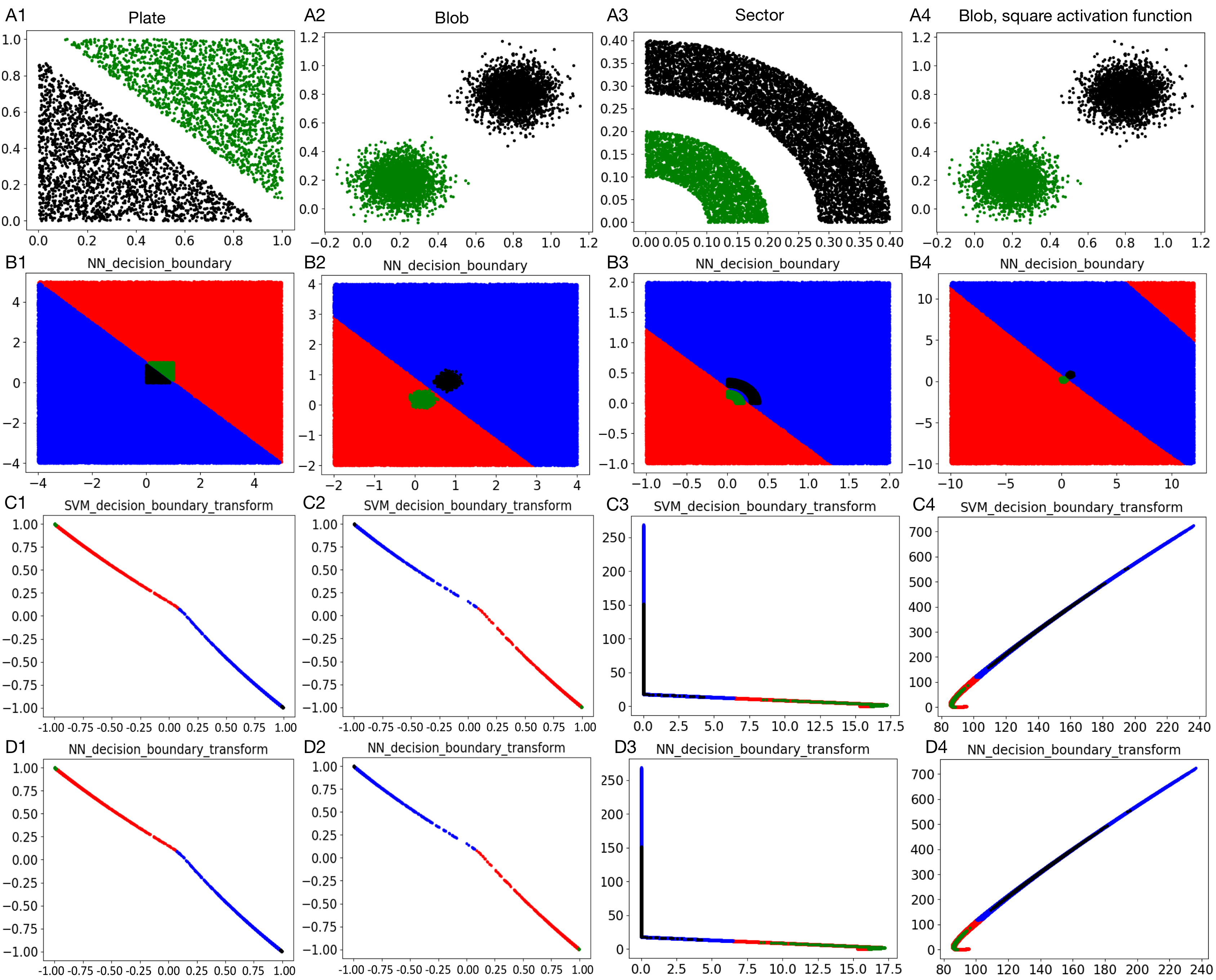}}
  \caption{Additional results of the simulated data, whose data range is much small than that in the main text. The four rows have the same meanings as the figure in the main text. The first three columns are the results of the neural networks with ReLU activation function on three linearly separable datasets. The last column is the result of the neural network with square activation function trained on Blob dataset.}
  \label{fig:simulated_small}
\end{figure*}

\begin{figure*}[h!]
  \centerline{\includegraphics[width = 150mm]{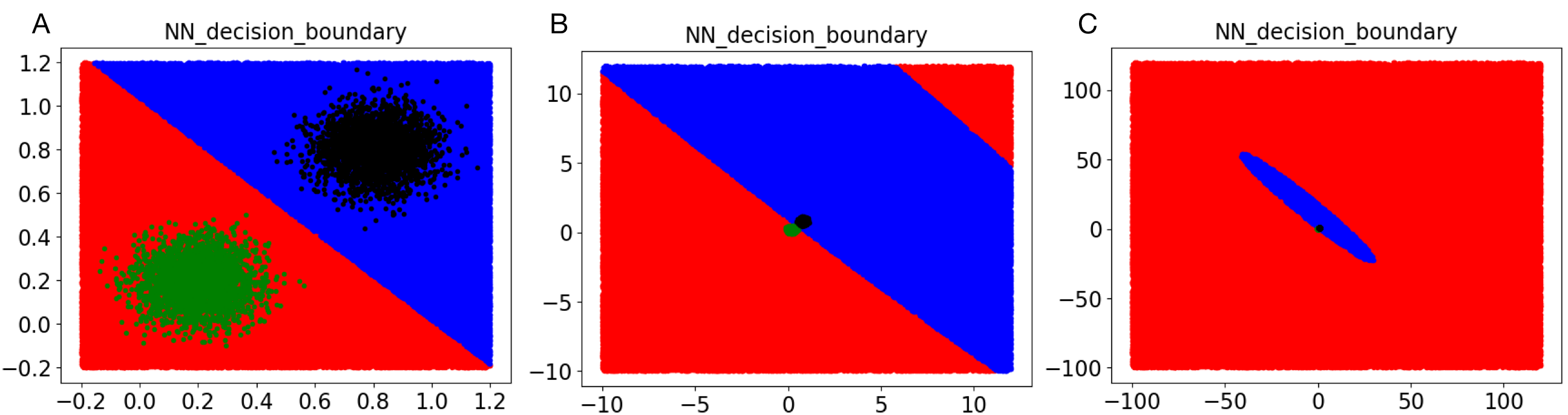}}
  \caption{The decision boundary of the neural network with square activation function trained on the Blob dataset in the original space in different scales.}
  \label{fig:square_result}
\end{figure*}

We first want to clarify why in the main text, we chose the data range of the simulated linearly separable datasets to be relatively large, from $0$ to around $100$ or $400$. Here we provide the results of the small-range linearly separable datasets (from $0$ to around $1$), which can be referred to Fig. \ref{fig:simulated_small}. Intuitively, the decision boundary in the original space (Fig. \ref{fig:simulated_small} (B1-B3)) is very surprising because the highly over-parameterized multi-layer neural network seems to learn a linear decision boundary. We argue that it is because of the small range of the datasets and also the shape of the activation function. As we know, a very large part of the ReLU activation function is linear. If the data range is very small, it is very likely that during training, the nonlinear part of the activation function is not used. As a result, the whole network becomes a linear classifier, which makes the decision boundary linear. We demonstrate that by performing an additional experiment on the small-range Blob dataset with the neural network having the following square activation function:
\begin{gather*}
    a(u) = u^2.
\end{gather*}
Within this function, there is no linear part. So even the data range is small, the decision boundary of the neural network should still be nonlinear. The experimental results of this setting are shown in the last column of Fig. \ref{fig:simulated_small}. From Fig. \ref{fig:simulated_small} (B4), we can see that the decision boundary in the original space is a nonlinear one, which is as expected. On the other hand, we also show the decision boundary in the original space in different scales in Fig. \ref{fig:square_result}. As shown in Fig. \ref{fig:square_result} (A, B), although the boundary is nonlinear globally (Fig. \ref{fig:square_result} (C)), it is very similar to a linear boundary if we only consider its local shape (i.e. from $-1$ to $2$), which supports our assumption, that is, if the data range is small, the nonlinear power of the activation function is used limitedly. This experiment demonstrates that the data range combining with the activation function can have a significant impact on the decision boundary in the original space. To eliminate the potential misunderstanding and misleading results caused by the datasets and emphasize the main results, we chose the large-range datasets in the main text. 

On the other hand, if we investigate the results of the neural network (Fig. \ref{fig:simulated_small} (D1-D4)) and the linear SVM (Fig. \ref{fig:simulated_small} (C1-C4)) in the tranformed space on those small-range datasets, we can find that the results are similar to those on the large-range datasets in the main text, which further supports our main results.

\end{appendices}

\end{document}